%% file: mm2021.tex
\documentclass[sigconf]{acmart}
\usepackage[ruled,linesnumbered]{algorithm2e}
\usepackage{anyfontsize}
\usepackage{bbding}
\usepackage{subfigure}
\usepackage{textcomp}
\usepackage{color}
\usepackage{balance}

\AtBeginDocument{%
  \providecommand\BibTeX{{%
    \normalfont B\kern-0.5em{\scshape i\kern-0.25em b}\kern-0.8em\TeX}}}

\copyrightyear{2021}
\acmYear{2021}
\setcopyright{acmcopyright}
\acmDOI{10.1145/3474085.3475530}

\acmConference[MM '21]{Proceedings of the 29th ACM International Conference on Multimedia}{October 20--24, 2021}{Virtual Event, China}
\acmBooktitle{Proceedings of the 29th ACM International Conference on Multimedia (MM '21), October 20--24, 2021, Virtual Event, China}
\acmPrice{15.00}
\acmISBN{978-1-4503-8651-7/21/10}

\acmSubmissionID{1932}

\begin{document}
\fancyhead{}
\sloppy

\title{Towards Accurate Localization by Instance Search}


\author{Yi-Geng Hong}
\affiliation{%
 \institution{Xiamen University}
 \city{Xiamen}
 \country{China}
}
\email{yghong@stu.xmu.edu.cn}

\author{Hui-Chu Xiao}
\affiliation{%
 \institution{Xiamen University}
 \city{Xiamen}
 \country{China}
}
\email{hcxiao@stu.xmu.edu.cn}

\author{Wan-Lei Zhao}
\authornote{Corresponding author.}
\affiliation{%
 \institution{Xiamen University}
 \city{Xiamen}
 \country{China}
}
\email{wlzhao@xmu.edu.cn}

\renewcommand{\shortauthors}{Hong, et al.}

\begin{abstract}
Visual object localization is the key step in a series of object detection tasks. In the literature, high localization accuracy is achieved with the mainstream strongly supervised frameworks. However, such methods require object-level annotations and are unable to detect objects of unknown categories. Weakly supervised methods face similar difficulties. In this paper, a self-paced learning framework is proposed to achieve accurate object localization on the rank list returned by instance search. The proposed framework mines the target instance gradually from the queries and their corresponding top-ranked search results. Since a common instance is shared between the query and the images in the rank list, the target visual instance can be accurately localized even without knowing what the object category is. In addition to performing localization on instance search, the issue of few-shot object detection is also addressed under the same framework. Superior performance over state-of-the-art methods is observed on both tasks.
\end{abstract}



\maketitle

\input{intr}
\input{relat}

\input{method}
\input{exp}
\input{conclusion}

\begin{acks}
  This work is supported by National Natural Science Foundation of China under grants 61572408 and 61972326.
\end{acks}

\bibliographystyle{ACM-Reference-Format}
\balance
\bibliography{ref}

\end{document}

%% file: intr.tex
\section{Introduction}
Visual object localization from image/video frame is a crucial step in object detection~\cite{FasterRCNN, YOLO}, visual object tracking~\cite{SiamFC, SiamRPN, SiamRPN++}, as well as instance search~\cite{FCIS+XD, PCL+SPN, DASR}, etc. It requires to outline the position (usually given as a bounding box) where the target object is located. In object detection, this problem has been well-addressed by several mainstream detection frameworks such as Faster R-CNN~\cite{FasterRCNN} and YOLO~\cite{YOLO}, while most of which require object-level annotations for the training images. When there are only the image-level class labels available, object detection becomes challenging. This issue is widely known as weakly supervised object detection (WSOD)~\cite{WSDDN, PCL, SelfPacedPAMI2019}. It becomes even more demanding when only a few annotations are available for each object category, which is known as few-shot object detection~\cite{MetaR-CNN, Meta-RCNN, FewX, FeatureReweighting}. Single visual object tracking and instance search face a similar situation as few-shot object detection. In both of them, the target instance is required to be localized with the given instance from only one image/shot. Instance search differs from visual object tracking in the sense that no temporal clues could be capitalized, making the localization even harder.

\begin{figure}[t]
  \centering
  \includegraphics[width=\linewidth]{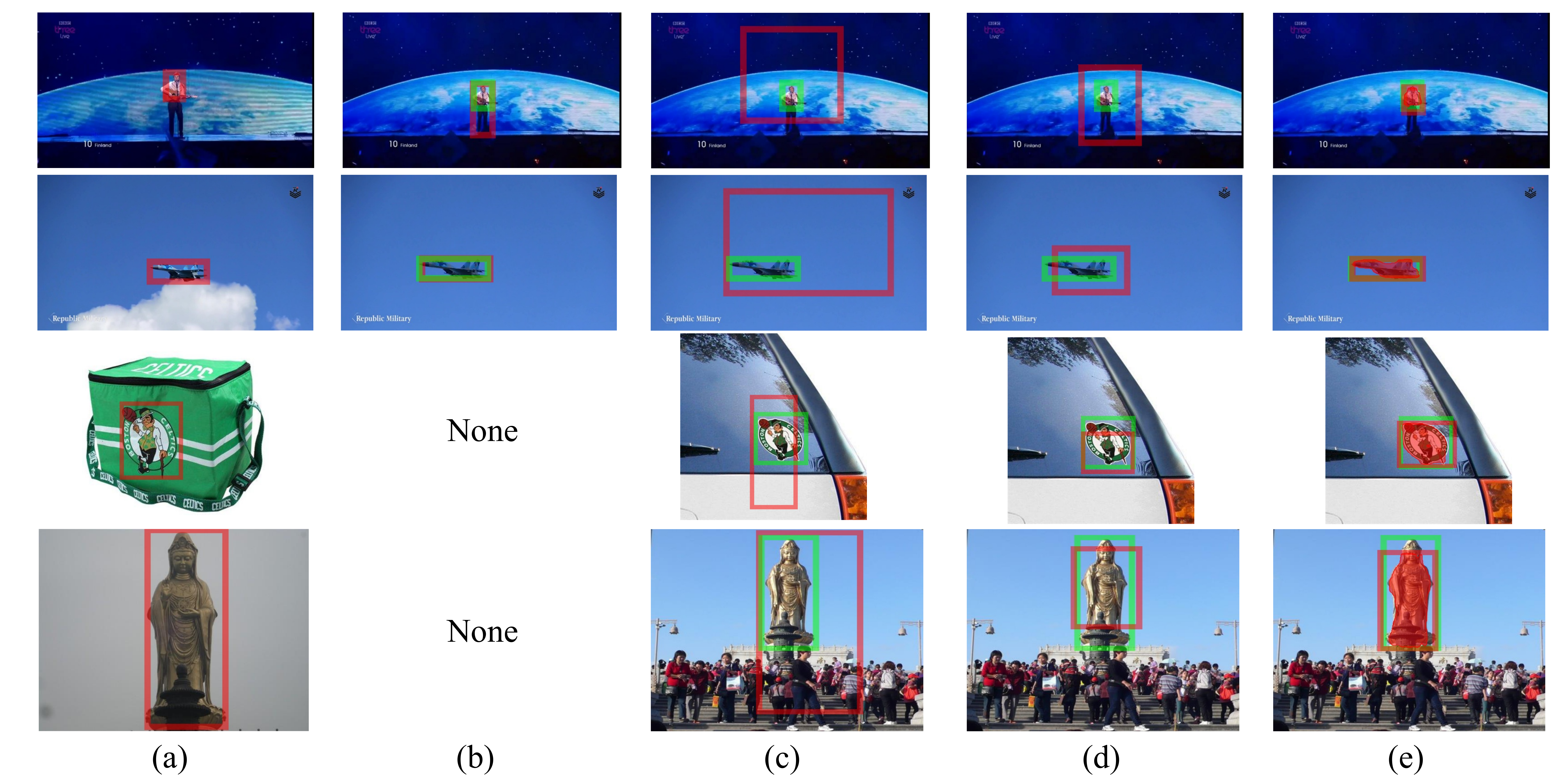}
  \Description{Instance localization results by different instance search methods.}
  \caption{Illustration of instance localization results from four instance search methods. The ground-truth boxes are in green and the produced boxes from different methods are in red. (a): Query, (b): Strongly supervised, (c): Weakly supervised, (d): Unsupervised, (e): Ours.}
  \label{fig:example}
\end{figure}

In instance search, the search system is required to return a rank list of images where the query instance appears. In the meantime, the location of the target instance in these images should be outlined. The localization issue has been addressed following different models that are designed originally for object detection. Early attempt~\cite{DeepVision} is based on strongly supervised models. Although superior performance is reported, it is only effective when object-level annotations are available. This issue has been alleviated by using a weakly supervised object detection framework. In~\cite{PCL+SPN}, the regions which have high class responses are detected as objects. Whereas such kind of method only works on the known object categories. In practice, no restriction should be made on the category of a supplied query instance from the end-user. As illustrated in \autoref{fig:example}(b)-(c), the localization is not achievable for the strongly supervised method on the unknown instances, and the performance from the weakly supervised method is far from satisfactory.

Recent research~\cite{DASR} shows that deeply activated regions that are output by a pre-trained image classification network for an input image coincide well with the instances inside the image. Features derived from these regions demonstrate good performance in instance search task. Additionally, this method is able to detect objects from both the known and unknown categories since these regions are localized by backpropagating from the class-agnostic layer~\cite{DASR}. Unfortunately, only several sub-regions of an object are activated to support the classification. As a result, the localization is only able to cover part of the object, which is illustrated in \autoref{fig:example}(d).

In this paper, we aim to achieve accurate object localization on the top-ranked images returned by instance search~\cite{DASR}. Although it sounds like a post-processing step for instance search task, it turns out to be a novel way to address the detection issue in few-shot setting. Compared with performing localization during the search, localization on the top-ranked images is easier to undertake since a common instance is shared between most of the images in the list and the query. In this scenario, the target instance from all the images can be easily outlined even without knowing what the object category is. Under this principle, an iterative self-paced training procedure is designed to fulfill the localization task. Given the initial locations returned by instance search~\cite{DASR} and the query instance, the locations of the target instance are incrementally refined as the training proceeds. \autoref{fig:example}(e) illustrates the performance of our method on both known and unknown object categories. The contributions of this paper are three folds.

\begin{itemize}
	\item {First of all, an iterative self-paced training procedure is proposed. Inside one round of iteration, the instance regions are activated by a Siamese network in the candidate images by referring to the query instance. The bounding box is generated by the standard Mask R-CNN with the correlation feature map output from the Siamese network, and refined during the iterative self-paced training;}
	\item {Additionally, a contrastive training strategy is used to enhance the discriminativeness of the network, and a co-attention module is employed to produce pseudo ground-truth mask for Mask R-CNN. Both of them lead to higher localization accuracy;}
	\item {The proposed framework is not only effective in instance search, but a novel solution for few-shot object detection.}
\end{itemize}

As will be revealed later, the localization accuracy of this method could be even better than the strongly supervised methods. In addition to instance localization, we also show how the few-shot object detection can be addressed with the proposed framework.

The rest of this paper is organized as follows. Section \ref{sec:related-work} reviews relevant research works about instance search, weakly supervised and few-shot object detection. Our method is presented in Section \ref{sec:method}. Section \ref{sec:apps} shows a practical way of our method to address the few-shot object detection. The empirical studies about our method are presented in Section \ref{sec:experiments}. Section \ref{sec:conclusion} concludes this work.

%% file: relat.tex
\section{Related Work}
\label{sec:related-work}
In this paper, we are going to address the object localization issue from the perspective of instance search, where only one shot is used for supervision. For this reason, the reviews in this section focus on relevant works in instance search and object detection tasks that follow either weakly supervised or few-shot paradigms.

\subsection{Instance Search}
Conventionally, instance search is addressed as an image search task~\cite{R-MAC, CroW, class-weighted, BLCF, RegionalAttention}. Since they are unable to localize instances, only a global feature is extracted for the whole image. During the feature extraction, higher weights are assigned to the potential instance regions. The feature is usually not discriminative since it is a mixture of features from several latent instances. In a recent method~\cite{LocalSimilarity}, the similarity between the query instance and the candidate images is learned via a ConvNet. Since the network is able to learn the spatial correlation between the regions in the candidate image and the query instance, high similarity is returned when the target instance appears in the image. Superior performance is reported on several benchmarks. However, the instance search is essentially performed via a deep convolution between the query instance and all the candidate images. It is hardly scalable to million-level due to the high time costs.

Recently, several methods~\cite{DeepVision, PCL+SPN, DASR} based on instance-level feature representation are proposed. Although they are designed under different training conditions, the frameworks are generally similar. There are two major steps in these methods, namely instance localization and feature extraction from the detected instance regions. Among these methods, deeply activated salient region (DASR)~\cite{DASR} relies only on a pre-trained image classification network. It shows much stable search performance over~\cite{DeepVision, PCL+SPN} and remains effective for unknown categories. However, the localization only covers a part of an object.

\begin{figure*}[t]
  \centering
  \includegraphics[width=0.85\linewidth]{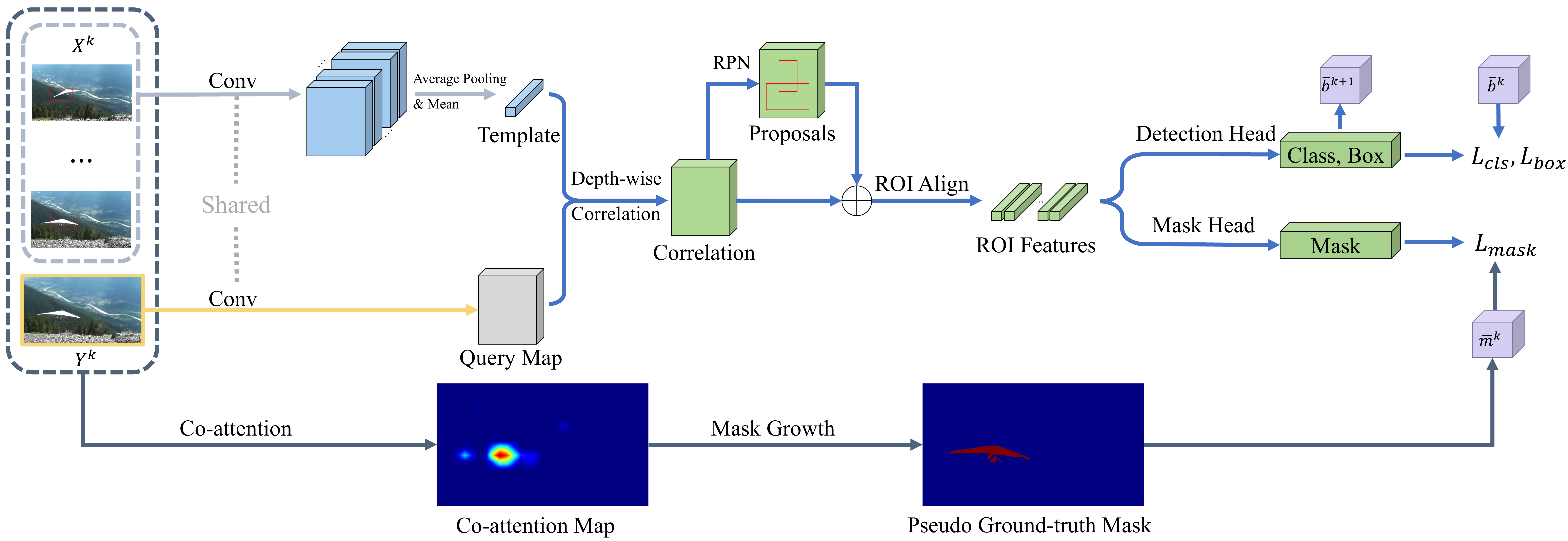}
  \Description{Architecture of self-paced instance localization.}
  \caption{Architecture of self-paced instance localization. The localization is trained on the Y-branch of a Siamese network. Feature maps are produced for $Y^k$ by correlating with features of $X^k$. Standard Mask R-CNN localizes the target instance from the correlation feature maps. The co-attention module produces pseudo masks to support the training of Mask R-CNN.}
  \label{fig:overview}
\end{figure*}

\subsection{Weakly Supervised Object Detection}
Weakly supervised object detection addresses the localization issue under the restriction that only image-level annotations are available. There are several popular methods~\cite{ren2015weakly, cinbis2016weakly, WSDDN, PCL} in recent literature. They all follow the pipeline of multiple instance learning (MIL)~\cite{maron1998framework}. Specifically, an image is taken as a bag of proposals. Each proposal is fed to the networks to judge whether an object is in presence.

Since only image-level annotations are available, it becomes a common practice \cite{PCL,SelfPacedPAMI2019, C-SPCL, high-quality-proposals, improving-detection} to generate pseudo ground-truth boxes for training. In PCL~\cite{PCL}, a cascade of multiple instance classifiers is used. The classification score that one instance classifier outputs is used as the supervision of the following instance classifier. Sangineto \emph{et al.} \cite{SelfPacedPAMI2019} propose a self-paced learning paradigm to select the most reliable pseudo ground-truth boxes for training iteratively. Zhang \emph{et al.} \cite{C-SPCL} obtain pseudo ground-truth boxes by judging the simplicity of both the image and the instance. In the methods above, the reliability of the generated pseudo ground-truth boxes is measured by the confidence score from the category classifier. As a result, the selected pseudo ground-truth boxes usually cover only the most discriminative part of the object since they contribute the most to classification. 

\subsection{Few-Shot Object Detection}
Few-shot object detection aims to detect objects of novel categories with very few training samples. As several aspects are shared in common between single object tracking and one-shot object detection, several works~\cite{FeatureReweighting, MetaR-CNN, CoAE, FewX} adopt the Siamese architecture, which has been proven to be effective in object tracking~\cite{SiamRPN, SiamRPN++, DaSiamRPN}, for few-shot object detection. In these methods, the training samples of each new category are used to generate category-specific information, which acts as the templates. Specifically, MetaYOLO \cite{FeatureReweighting} and Meta R-CNN \cite{MetaR-CNN} generate re-weighting vectors using the few samples for each novel category. These re-weighting vectors are in turn used to re-weight the channels of query feature maps. CoAE \cite{CoAE} uses non-local block \cite{NonLocal} and squeeze-and-excitation \cite{SENet} module to build the correlation between the query and target. FSOD \cite{FewX} uses the few samples to generate a convolution kernel. The depth-wise convolution is performed on the query feature map with the convolution kernel. The object regions that are correlated to the training samples are activated.

In the above detection frameworks, Siamese networks \cite{SiameseOrigin} play the key role. Essentially, it is a weight-shared neural network that has two input branches. Such structure makes it suitable for the tasks that require comparison or matching. A detector that adopts the Siamese architecture detects objects in the query branch by referring to the template branch. Since no category label is required in training, it is able to detect objects of both known and unknown categories as long as a correct template is provided.

In this paper, similar to few-shot detection, the Siamese network is adopted to learn the correlation between the query instance and the candidate images. High responses are generated from the candidate image regions where the target instance localizes. The correlation is fed into standard Mask R-CNN to localize the target instance.

%% file: method.tex
\section{Self-Paced Instance Localization}
\label{sec:method}
As discussed in Section~\ref{sec:related-work}, the object localization with the existing methods is either unachievable or imprecise due to the limited supervision in few-shot learning and weakly supervised learning. In this paper, we address this issue from the perspective of instance search. Given a list of top-ranked images are returned for a specified query, a common instance is shared between the query and most of the candidate images assuming that the search quality is sufficiently high (i.e., mAP $\geq$ 0.5). The localization of an instance from all these images can be largely viewed as a common object mining task. Compared with the method in~\cite{LocalSimilarity}, the cost of this mining task is much lower, given the fact that only very few images need to be processed. Compared with the methods in~\cite{CoAE,SENet}, more potentially positive training samples are available from the rank list, making the training much easier to undertake. 

Our method works under the following two assumptions. 1. The bounding box provided by the user covers a semantically meaningful instance; 2. Most of the top rank list images are true positives, which is achievable with the recent method~\cite{DASR}\footnote{There is no preference over the instance search method as long as it supplies the candidate images of good quality and remains sensitive to unknown object categories.}. In order to facilitate our discussion in this section, several symbols are defined as follows. In our discussion, $Q=\{q_i\}_{i=1}^{N}$ is the set of the query images. Correspondingly, the bounding boxes to outline the query instances are denoted by $U=\{u_i\}_{i=1}^{N}$, where $u_i$ is the box of $q_i$. $R \in \mathbb{R}^{N \times M}$ are the top-$M$ candidate images for the $N$ queries. $Z \in \mathbb{R}^{N \times M \times 4}$ are the corresponding bounding boxes of $R$. Specifically, $R_{ij}$ and $Z_{ij}$ are respectively the image and box of the $j$-th candidate of the $i$-th query.

\subsection{Localization Network Architecture}
\label{sec:network}
The overall architecture of mining target instances from the top-ranked images is shown in \autoref{fig:overview}. There are three major components, namely the Siamese network, Mask R-CNN~\cite{mask-rcnn}, and the co-attention module. The query image and the candidate images are input into two branches of the Siamese network. The instance regions are activated in the candidate images by correlating with the query instance. The produced correlation feature map by the Siamese network is fed to Mask R-CNN to localize the target instance in the image. Given the $i$-th proposal produced by RPN, the detection head predicts the probability $p_i$ and box $b_i$, while the mask head predicts the object mask $m_i$. Since our primary goal is to localize the target instance, the mask head is introduced mainly to enhance the localization accuracy. The co-attention module produces pseudo ground-truth mask for Mask R-CNN, which will be detailed in Section \ref{sec:co-attention}.

Specifically, the localization issue in instance search is addressed as a self-paced multi-task learning consisting of multiple stages. The loss function at stage-$k$ is defined as
\begin{equation}
\begin{aligned}
  L_{total}^k = \sum_{i} \omega_i^k \{L_{cls}^k(p_i^k, \bar{p_i}^k) + L_{box}^k(b_i^k, \bar{b_i}^k) + L_{mask}^k(m_i^k, \bar{m_i}^k)\},  \label{eqn:loss1}
\end{aligned}
\end{equation}
where $\omega_i^k$ is a weight factor and $\bar{p_i}^k, \bar{b_i}^k, \bar{m_i}^k$ are ground-truth labels. Because only the ground-truth boxes of query images $U$ are available at the beginning, the model can be well-trained only after several self-paced training stages. That is, the ground-truth boxes $\bar{b_i}^k$ along with $\omega_i^k$ for the candidate images are derived from the previous training stage. The classification loss $L_{cls}$, the bounding box loss $L_{box}$, and the mask loss $L_{mask}$ are defined the same as standard Mask R-CNN \cite{mask-rcnn}. At the beginning, only box for the query image is available, while the ground-truths for all the candidate images are missing. At this stage (namely Stage-0), the candidate images are input to the X-branch of the Siamese network and the query image is input to the Y-branch of the network. The above loss function at Stage-0 is rewritten as
  \begin{equation}
  \begin{aligned}
    L_{total}^0 = \sum_{i} \{L_{cls}(p_i^0, \bar{p_i}^0) + L_{box}(b_i^0, u_i) + L_{mask}(m_i^0, \bar{m_i}^0)\}.  \label{eqn:loss2}
  \end{aligned}
  \end{equation}
In Equation~\ref{eqn:loss2}, the ground-truths $\bar{p_i}^0$, $u_i$, and $\bar{m_i}^0$ are all ready for the query image. The training is undertaken on the Y-branch and the trained weights will be shared with the X-branch.

After Stage-0, the trained network is able to localize target instance from the candidate images. The predicted boxes with high confidence are selected as pseudo ground-truth boxes for the next stage. The Y-branch now accepts not only the query image, but also the candidate images with pseudo ground-truth boxes. As the training continues, more and more candidates with high confidence boxes are joined as training data, which prevents the network from overfitting to query instances.

At Stage-$k$, a set of candidate instances $X^{k}$ and the query image $Y^{k}$ are input to the X-branch and the Y-branch, respectively. The features are extracted by a weight-shared extractor, namely $\mathcal{X}^{k} \in \mathbb{R}^{n \times s \times s \times c}$ for $X^{k}$, and $\mathcal{Y}^{k} \in \mathbb{R}^{1 \times h \times w \times c}$ for $Y^{k}$. The mean vector $\bar{\mathcal{X}}^k \in \mathbb{R}^{1 \times 1 \times 1 \times c}$ is given as

\begin{equation}
\bar{\mathcal{X}}^k = \frac{1}{n} \sum_{i=1}^n avg(\mathcal{X}_i), \label{eqn:xmean}
\end{equation}
where $avg(\cdot)$ is the spatial-wise average-pool function. $\bar{\mathcal{X}}^k$ is used as the kernel to slide on $\mathcal{Y}^{k}$ to compute the depth-wise cross correlation $\mathcal{C}^{k} \in \mathbb{R}^{1 \times h \times w \times c}$ \cite{SiamRPN++}. Thereafter, $\mathcal{C}^{k}$ is sent to Mask R-CNN. The detection and mask heads in Mask R-CNN predict probability $p^k$, boxes $b^k$, and masks $m^k$.

After one stage of training, the model is used to produce predictions for each candidate image. Among the produced boxes, the box with the highest probability from one image is selected as the pseudo ground-truth box, and its corresponding probability is used as the loss weight at next stage. Namely, we have

\begin{equation}
(\bar{b}^{k+1}, \omega^{k+1}) = \{(b^k_i, p^k_i) | \mathop{\arg\max}_{\{i | p^k_i > \tau^k\}} p^k_i\},
\label{}
\end{equation}
where $\tau^k$ is the confidence threshold at Stage-$k$. As the training continues, the threshold $\tau^k$ decreases since our model becomes more powerful and could deal with more difficult samples. $\omega^{k+1}$ is used as the weight factor in Equation~\ref{eqn:loss1} for the next stage.

In order to enhance the ability of discriminating unmatched pairs, a \textit{contrastive training} strategy is adopted to help our network to distinguish different instances. Namely, candidates $X_p^{k}$ that the target instance appears are coupled with $Y^{k}$ as the positive pairs. Another set of negative pairs are prepared by randomly sampling candidate instances $X_n^{k}$ from other queries in which the target instance does not appear. For the negative pairs, their ground-truth boxes and masks are always null. As a result, the network will be penalized for any predictions made on negative pairs.

\subsection{Co-attention Module}
\label{sec:co-attention}
In Equation~\ref{eqn:loss1}, $\bar{m}^{k}$ is the pseudo ground-truth mask for loss $L_{mask}^k(m_i^k, \bar{m_i}^k)$, and is generated by our proposed co-attention module. In the following, we show how $\bar{m}^{k}$ is calculated by the co-attention module. Since the generation procedure is identical for all stages, the superscript $k$ that marks the number of stage is omitted for simplicity.

Inspired by the deep descriptor transform (DDT) \cite{DDT}, the co-attention map is generated by applying PCA on the feature maps of query and its top-ranked images. Given a batch of $B$ images, we first compute their feature maps $\{\mathbf{X}_l\}_{l=1}^B$ using a pre-trained CNN feature extractor. Each feature map is an $H \times W \times C$ tensor. The mean vector is calculated from all spatial locations
\begin{equation}
\begin{aligned}
\bar{x} = \frac{1}{BHW} \sum_{l=1}^B \sum_{i=1}^H \sum_{j=1}^W \mathbf{X}_{l.i,j}. \label{eqn:featmap}
\end{aligned}
\end{equation}
The covariance matrix of $\bar{x}$ is given as
\begin{equation}
\begin{aligned}
Cov(x) = \frac{1}{BHW} \sum_{l=1}^B \sum_{i=1}^H \sum_{j=1}^W (\mathbf{X}_{l,i,j} - \bar{x})(\mathbf{X}_{l,i,j} - \bar{x})^T. \label{}
\end{aligned}
\end{equation}
The first principal component $\phi$ of matrix $Cov(x)$, which is the eigenvector corresponding to the greatest eigenvalue, is considered as the most common pattern across all the images. 

Given feature map of an input image is $F$, its corresponding co-attention map $A$ can be calculated by dot-product between $\phi$ and every spatial location $(i, j)$ on $F$, namely
\begin{equation}
\begin{aligned}
A_{i,j} = \phi^T \cdot (F_{i,j} - \bar{x}).  \label{}
\end{aligned}
\end{equation}
The element $A_{i,j}$ is large when the $F_{i,j}$ is similar with $\phi$, such that the most common pattern in the set of images are highlighted on the co-attention map $A$. The values in $A$ are normalized to $[0,1]$.

\begin{figure}[t]
  \centering
  \includegraphics[width=\linewidth]{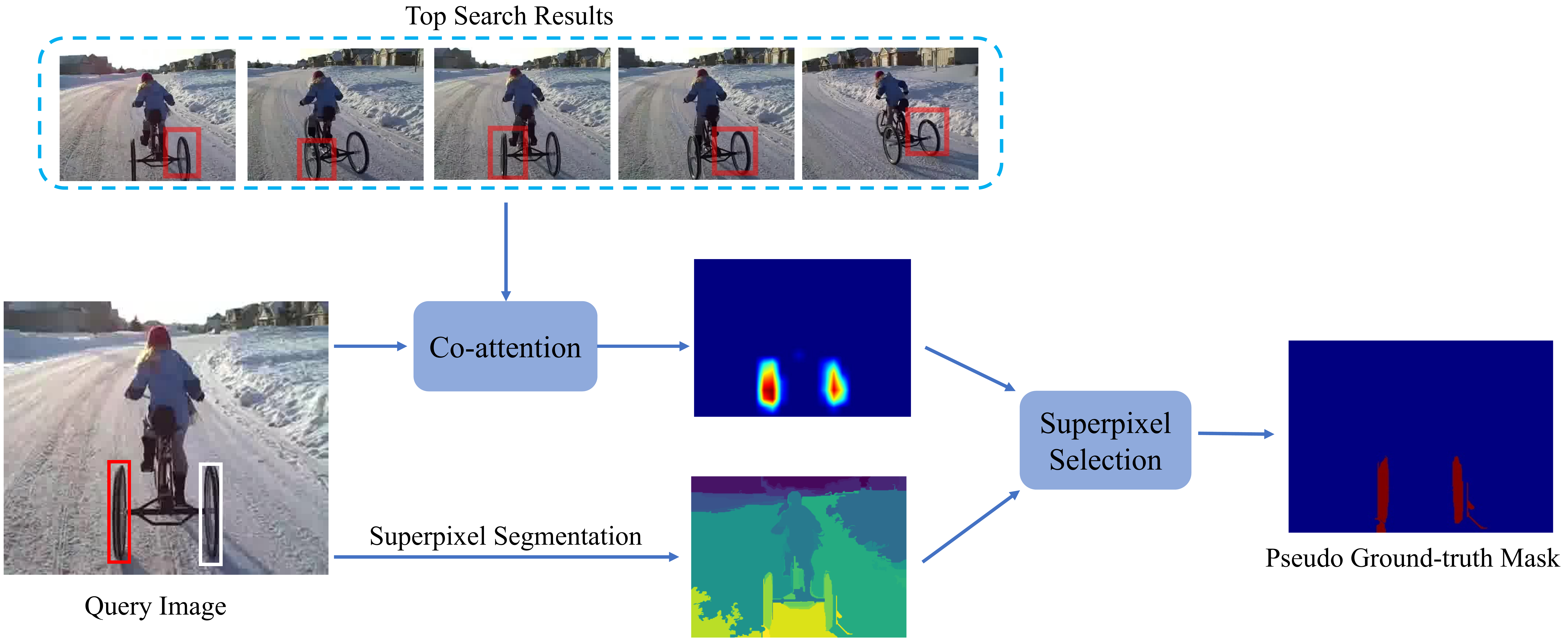}
  \Description{Instance localization results by different methods.}
  \caption{Illustration of pseudo ground-truth mask generation. The red box on the query image is the query box given by user, and the white box marks another instance that is unlabeled.}
  \label{fig:co-attention}
\end{figure}

\begin{algorithm}[t]
  \caption{Self-paced Training Procedure}
  \label{alg:training}
  \LinesNumbered
  \KwIn{Query images $Q$ and boxes $U$, top search result images $R$ and boxes $Z$, sample pool $V$, training stages $K$, iterations $P_k$ for stage $k$.}
  \KwOut{Trained localization network.}

  Initialize $V$ with $<R, Z>$ \\

  \For{$k=0:K$}{
    \For{$i=1:P_k$}{
      Select query from $<Q, U>$  as $<Y^k, \bar{b_i}^k>$ \\
      Sample from $V$ randomly as $X^k_p$ \\
      \If{k != 0}{
        Randomly exchange $<Y^k, \bar{b_i}^k>$ with one of $X^k_p$ \\
      }
      $\bar{m_i}^k \leftarrow$ \textit{Co-attention($X^k_p, Y^k$)} \\
      Select negative samples as $X^k_n$ from other rank list\\
      Update the network with loss function in Equation~\ref{eqn:loss1} \\
    }

    Update $V$ with confident predicted boxes \\
  }
\end{algorithm}

In this way, the pixels belonging to target instance are highlighted. However, directly taking $A$ as $\bar{m}$ would result in inferior performance since $A$ is too coarse as a mask. Instead, the mask $\bar{m}$ is produced by merging the superpixels~\cite{GraphSuperpixel} derived from the activated pixels. We compute a score $\theta_i$ for each superpixel $s_i$ by
\begin{equation}
\theta_i = \frac{\sum_{t \in s_i} A(t)}{|s_i|},  \label{eqn:phi}
\end{equation}
where $|s_i|$ is the number of pixels in $s_i$. Only the superpixels with scores above a threshold (\emph{e.g.}, $>$ 0.1) are selected. On the one hand, the co-attention module prepares the pseudo ground-truth mask $\bar{m}$ for training. On the other hand, it helps to discover the similar instances in the query, which could prevent the training from overfitting. As shown in \autoref{fig:co-attention}, the user launches a tyre as the query instance. However, there is another unlabeled tyre in the image. As they are similar to each other, the network could be confused and gets overfit with the specific instance. With the help of the co-attention map, the unlabeled similar instances are highlighted as positives, instead of being mistaken as false negatives.

\textbf{Summary on our SPIL framework} The detailed self-paced learning procedure is presented in Algorithm \ref{alg:training}. A sample pool $V$ is initialized with the query images and the corresponding top-rank instances (\textit{Line 1}). In one training iteration, a query instance is randomly sampled as $<Y^k, \bar{b_i}^k>$ and its five candidate instances are also randomly sampled from $V$ as $X_p^k$. $X_p^k$ and $Y^k$ are fed into the co-attention module to calculate the pseudo ground-truth mask $\bar{m}_i^k$. The query image $Y^k$ is input to the Y-branch of the network, while the candidate instances $X_p^k$ are input to the X-branch. To prepare for the contrastive training, five instances are sampled from another query as negative candidates $X_n^k$. The loss is computed and the network is updated (\textit{Lines 4--11}). The sample pool $V$ is updated with confident predicted boxes in the corresponding rank list (\textit{Line 13}). During the training, the query instance is swapped with a random instance from the five positive candidates (\textit{Lines 6--8}) except for Stage-0. Since our method is built based on self-paced training, it is now called self-paced training for instance localization (SPIL).

\section{Few-Shot Object Detection}
\label{sec:apps}
Few-shot object detection is largely similar to instance search in the sense that both of them require to localize the object from the image and they are trained under similar conditions. In this section, few-shot object detection is addressed from the perspective of instance search. Only a few labeled instances are provided for objects of novel categories. Since objects in one category are mostly similar in their appearance, the labeled and unlabeled objects of one category are close in the feature space. The few-shot detection by instance search is addressed in two steps. In the first step, all the annotated objects are treated as queries to query against all the unlabeled images. Usually, there are several objects in one image. In this case, each object in the image is treated as an independent query. The category labels of the query objects are therefore propagated from the  queries to the top-ranked images. In the second step, the proposed self-paced learning is applied to each query and the top-ranked list to localize the instance of a particular category. Following the same manner, more objects could be detected by selecting newly detected objects as queries, which is known as query expansion (QE).

In our implementation, we follow the challenging 10-shot 20-way setting to fulfill the few-shot object detection on the COCO dataset~\cite{COCO}. Namely, \textit{20} categories in Pascal VOC \cite{VOC} are chosen as novel categories, and the remaining \textit{60} categories as base categories. For each novel category, only \textit{10} labeled instances are provided. For each base category, all labeled instances are provided. The localization network presented in Section~\ref{sec:network} is first trained on all the base categories. The subsequent $K$ + 1 stages of training are carried out on novel categories in the same way as Section~\ref{sec:network}.

In order to boost the detection performance, a QE procedure is adopted. When the objects have been detected based on instance search, they are treated as query instances to search the image dataset again. These new objects used as queries will help to mine out more objects of the same category as they might be close to those undiscovered objects in the feature space. As will be revealed in the experiment, QE leads to another \textit{0.5\%} improvement.

%% file: exp.tex
\section{Experiments}
\label{sec:experiments}
In this section, the performance of the proposed method is studied on two tasks, namely, instance search and few-shot object detection.

\begin{table}[tb]
  \caption{Summary over datasets used in instance search task. Dataset Instance-251 is used for training, while datasets Instance-335 and INSTRE are used for testing}
  \label{tab:datasets}
    \begin{tabular}{lrr}
      \toprule
      Dataset & \#Query & \#Images   \\
      \midrule
      Instance-251 &  251 & 57,971 \\
      Instance-335 &  335 & 40,914  \\
      INSTRE &   1,250    & 27,293  \\   
      \bottomrule
    \end{tabular}
\end{table}

\subsection{Experimental Protocol}
The instance search performance is evaluated on two challenging datasets, namely Instance-335 \cite{DASR} and INSTRE \cite{INSTRE}. Instance-335 is derived from video sequences used in single visual object tracking~\cite{OTB, ALOV++, GOT10K, YoutubeBoundingBoxes, LaSOT}. INSTRE consists of \textit{28,543} images that cover \textit{250} different instances of diverse categories. Following the protocol in Iscen \emph{et al.} \cite{diffusionCVPR2017}, \textit{1,250} images are treated as queries (\textit{5} images for one instance), and the remaining \textit{27,293} images are treated as reference images. Another dataset Instance-251 originally used in single visual object tracking~\cite{GOT10K, YoutubeBoundingBoxes, LaSOT}, is collected for training. There is no intersection between Instance-251 and Instance-335. Details of the above datasets are shown in Table \ref{tab:datasets}. Only query instances and the corresponding search results are used during training.

In our implementation, DASR \cite{DASR} is called to produce the top search results. Our method is adopted to localize target instances on the search results. Thereafter, the boxes produced by DASR are replaced by the boxes returned by our method. Accordingly, features produced by DASR are replaced by new features that are derived from the new boxes. The search results are therefore re-ranked with these new features. The mean Intersection over Union (mIoU) is adopted to evaluate the localization accuracy. The mean Average Precision (mAP) is adopted to measure the search performance. The performance of our method is studied in comparison to instance-level features based methods, such as FCIS+XD~\cite{FCIS+XD}, PCL*+SPN \cite{PCL+SPN}, and DASR~\cite{DASR}. All of them are capable of localizing objects. Image-level features based methods \cite{R-MAC, CroW, class-weighted, BLCF,RegionalAttention, DeepVision} are also compared. Since FCIS+XD~\cite{FCIS+XD} is a strongly supervised method trained on the COCO dataset, it is only effective on instances belonging to the \textit{80} categories of COCO.

For few-shot object detection, we follow the challenging 10-shot 20-way evaluation settings on COCO \cite{COCO}. The Average Precision (AP) is adopted as the evaluation measure. The performance of our method is studied in comparison to state-of-the-art methods \cite{FeatureReweighting, MetaDet,MetaR-CNN, TFA, Meta-RCNN, MPSR, FewX, SRR-FSD, FSCE}.

\textit{Implementation details.} ResNet-50 \cite{ResNet} pre-trained on ImageNet \cite{ImageNet} is adopted as the backbone and the feature extractor in our framework. The number of top search results is fixed to \textit{128}. The network is trained for \textit{4} stages, namely $k=[0,~3]$. We set the confidence threshold $\tau^k=0.99-0.1 \times (k-1), k=\{1,2,3\}$. The superpixel merge threshold is fixed to $0.1$. The images input to the X-branch are cropped around the instance and then resized to $320 \times 320$. The images input to the Y-branch are resized with the shorter side to \textit{600} pixels and the longer side to \textit{1000} pixels. For a query image, the five positive training pairs are the coupling of the query with five images from its top-128 rank list. The five negative training pairs are sampled from top-ranked images of another query. Our network is trained on \textit{4} Nvidia 1080Ti GPUs using SGD with a batch size of \textit{8}. At Stage-0, the learning rate is \textit{0.001} for the first \textit{12,000} iterations and \textit{0.0001} for later \textit{3,000} iterations. At Stage-$k$ $(k > 0)$, the learning rate is \textit{0.001} for the first \textit{7,000} iterations and \textit{0.0001} for later \textit{3,000} iterations.

\begin{figure}[t]
  \centering
  \includegraphics[width=\linewidth]{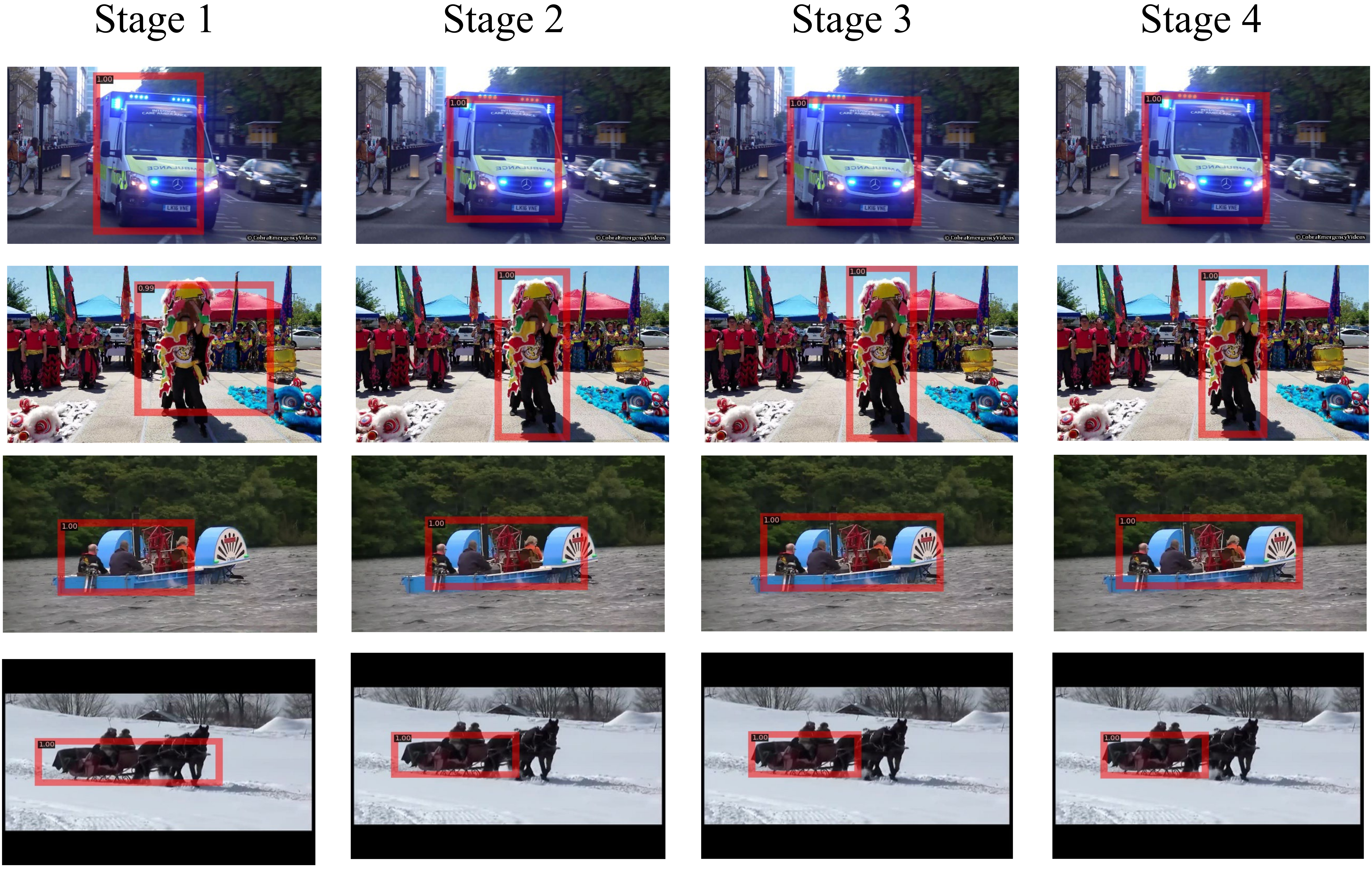}
  \caption{The variation of pseudo ground-truth boxes on four training stages.}
  \label{fig:progress-example}
\end{figure}

\begin{table}[tb]
  \caption{Ablation study over proposed components on Instance-335 and INSTRE. \emph{SP}: self-paced training strategy. \emph{CA}: co-attention module. \emph{CT}: contrastive training strategy}
  \label{tab:ablation}
    \begin{tabular}{lcc}
      \toprule
      Method & Instance-335 & INSTRE   \\
      \midrule
      DASR~\cite{DASR} &   27.03   &  34.98  \\
      One-shot FSOD~\cite{FewX} &   41.98   &  42.52  \\ \hline     
      SP         &  40.57 & 43.48 \\
      SP+CA      &  40.90 & 43.83 \\
      SP+CT      &  46.32 &  52.56 \\
      SP+CT+CA   &  \textbf{47.45} & \textbf{53.92} \\
      \bottomrule
    \end{tabular}
\end{table}

\begin{figure*}[htp]
  \centering
  \includegraphics[width=\linewidth]{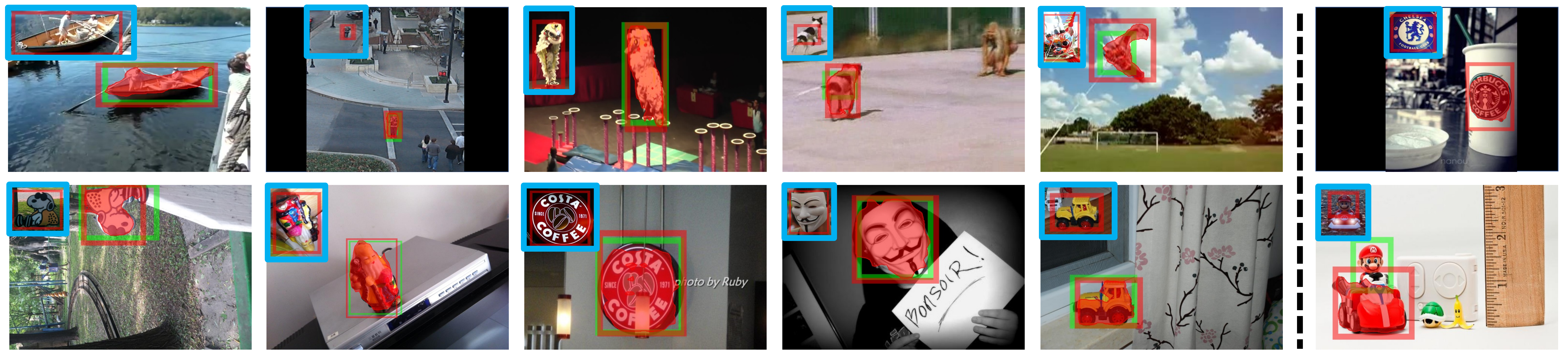}
  \Description{Qualitative results on Instance-335 and INSTRE.}
  \caption{Qualitative results on Instance-335 and INSTRE. Query images (blue border) are placed in the upper left of the image. The first to the fifth columns present successful cases and the last column presents failed cases. The detection results are shown with red bounding boxes and masks, while the ground-truths are shown with green bounding boxes.}
  \label{fig:more-examples}
\end{figure*}

\subsection{Ablation Study}
In this section, an ablation study is carried out to investigate the contributions of the major components in our method, namely the \textit{self-paced training}, \textit{co-attention} module, and the \textit{contrastive training} strategy. The \textit{self-paced training} is proposed to gradually mine useful knowledge from the search results. Without the self-paced training strategy, the problem is reduced to one-shot object detection. The \textit{co-attention} module supplies the pseudo ground-truth masks for Mask R-CNN by mining similar instance regions on the image. The \textit{contrastive training} strategy enhances discriminativeness of the network to positive pairs over negative pairs.

Four runs of our method are carried out under different configurations. In the first run (SP), the self-paced training is carried out without the co-attention module and contrastive training. More and more candidate images with pseudo ground-truth boxes are gradually joined in the Y-branch. It is comparable to few-shot object detection method \cite{FewX} in one-shot setting. In the second run (SP+CA), the self-paced training is undertaken with the assistance of the co-attention module. In the third run (SP+CT), contrastive training is employed, which is expected to enhance the discriminativeness of the network. In the fourth run, both the co-attention module and the contrastive training strategy are adopted in the self-paced training. In the evaluation, the performance from FSOD under one-shot setting~\cite{FewX} and DASR~\cite{DASR} is treated as the comparison baseline. For one-shot FSOD, it is similar to the setting without ``SP'' as both are trained with only query instances. In this experiment, we mainly study the localization accuracy (mIoU) of the considered methods. The results are presented in \autoref{tab:ablation}.

As shown in the table, the localization accuracy of ``SP'' is much higher than DASR, which shows the effectiveness of the self-paced training strategy. ``SP'' performs similarly as one-shot FSOD when no other strategies are integrated with self-paced training. Considerable improvement is observed when contrastive training is integrated with self-paced training, which could enhance the discriminativeness of the network. The integration of the co-attention module leads to an extra \textit{1.2\%} improvement. \autoref{fig:progress-example} shows the variations of localization boxes on three instances as the self-paced training proceeds from Stage-1 to Stage-4. Most of the boxes become precise at Stage-3, while the refinement tends to converge after this stage. Since self-paced training, contrastive training, as well as the co-attention module are all helpful for the localization, all of them are integrated into our method in the following experiments.

\subsection{SPIL on Instance Search}

\begin{table}[tb]
  \caption{Comparison on localization accuracy (mIoU) on Instance-335 and INSTRE. Top-128 rank list is considered for each query}
  \label{tab:miou}
  
  \begin{tabular}{lccc}
    \toprule
    Method  & Extra data & Instance-335  & INSTRE     \\
    \midrule
    FCIS+XD \cite{FCIS+XD}  & COCO & \textbf{51.00}  & - \\
    PCL*+SPN \cite{PCL+SPN} & COCO & 23.10 & 38.80 \\
    DASR    \cite{DASR}     & None                  & 27.03 & 34.98 \\
    \midrule
    SPIL   & None & 47.45 & \textbf{53.92} \\
    \bottomrule
  \end{tabular}
  
\end{table}

In this section, the performance of our method is studied in instance search in comparison to representative works in the state-of-the-arts. We mainly consider the localization accuracy as well as the search quality. For instance localization, all three methods in the literature that are capable of localizing instances are considered. They are FCIS+XD~\cite{FCIS+XD}, PCL*+SPN~\cite{PCL+SPN}, and DASR~\cite{DASR}. Among these methods, FCIS+XD is strongly supervised and requires pixel-level annotations. It only works for instances that fall in \textit{80} COCO \cite{COCO} categories. While for PCL*+SPN, image-level annotations are required. Our method SPIL is built upon DASR, both of which are based on pre-trained image classification CNN. No extra training on annotated data is required. Note that no annotation is required for the training on Instance-251 for SPIL.

\begin{table}[]
  \caption{Search quality (mAP) on Instance-335 and INSTRE. Top-128 rank list is considered for each query}
  \label{tab:mAP}
  \begin{tabular}{lcc}
    \toprule
    Method  & Instance-335  & INSTRE   \\
    \midrule
    R-MAC \cite{R-MAC}      & 31.98   & 39.81 \\
    CroW  \cite{CroW}       & 33.90   & 43.80 \\
    CAM   \cite{class-weighted} & 28.31 & 23.97 \\
    BLCF-SalGAN \cite{BLCF} & 38.16   & 48.88 \\
    RegionalAttention \cite{RegionalAttention}  & 38.53 & 38.00 \\
    DeepVision  \cite{DeepVision} & 56.05 & 6.20 \\
    FCIS+XD \cite{FCIS+XD}  & 52.26   & - \\
    PCL*+SPN \cite{PCL+SPN} & 50.01   & 43.00 \\
    \midrule
    DASR    \cite{DASR}     & 59.73   & 45.37 \\
    DASR+SPIL               & \textbf{59.74}   & \textbf{45.49}\\
    \bottomrule
  \end{tabular}
\end{table}

The mIoUs from all four methods are calculated on top-128 search results and shown in Table~\ref{tab:miou}. For FCIS+XD, the localization on INSTRE~\cite{INSTRE} is not achievable as all the instances from INSTRE are unknown to it. As shown in the table, the mIoU from SPIL is comparable to FCIS+XD, which requires pixel-level supervision. At the same time, it maintains high performance on unknown object categories. SPIL outperforms the weakly supervised and unsupervised methods \cite{PCL+SPN, DASR} by a large margin.

\begin{figure}[t]
\begin{center}
  \subfigure[Instance-335]
      {\includegraphics[width=0.48\linewidth]{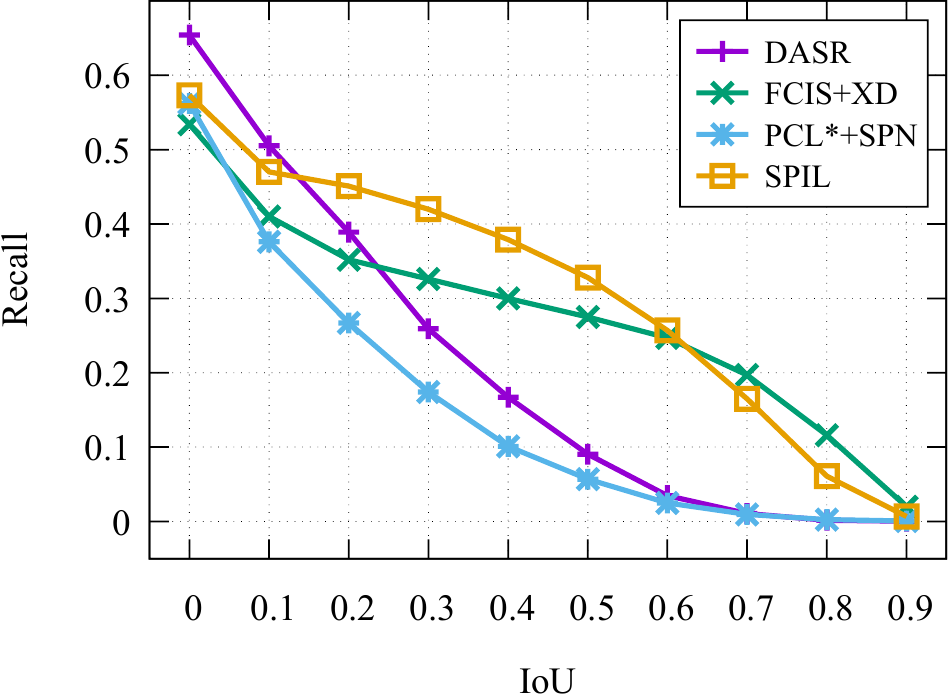}}
      \hspace{0.05in}
  \subfigure[INSTRE]
      {\includegraphics[width=0.48\linewidth]{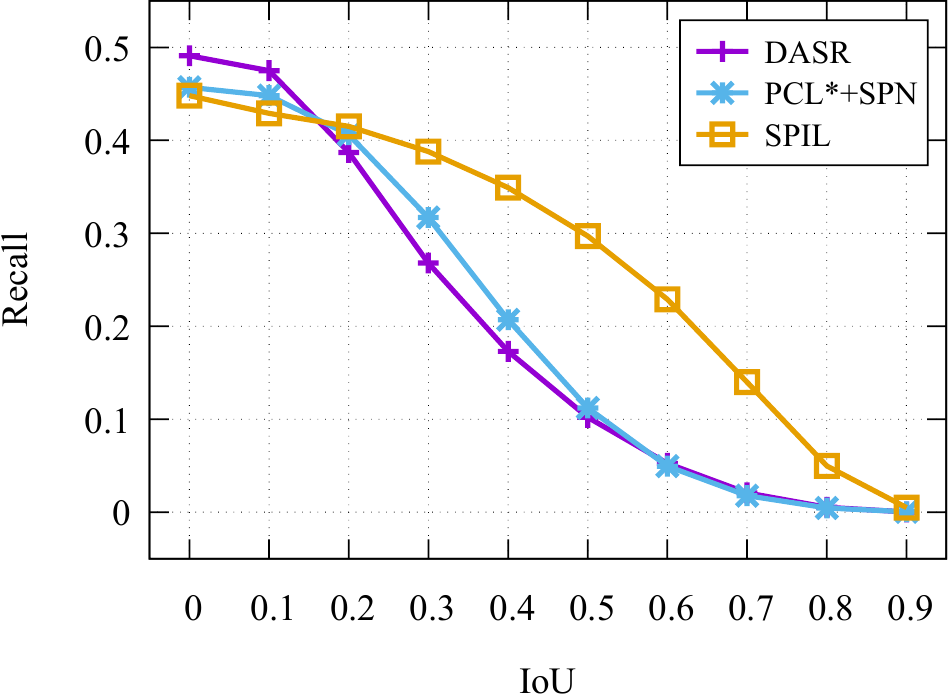}}
  \caption{Recall-IoU curves on Instance-335 and INSTRE.}
  \label{fig:recall-iou}
\end{center}
\end{figure}

The localization accuracy is further studied on the Recall-IoU curve. The recall is calculated under various IoU thresholds. \autoref{fig:recall-iou} shows the Recall-IoU curves of different methods on Instance-335 and INSTRE. As shown in \autoref{fig:recall-iou}(a), SPIL outperforms all the state-of-the-arts by a large margin, including FCIS+XD, which only shows higher performance when the threshold is high (\emph{e.g.}, \textit{0.7}).

When the precise boxes are produced for the candidate images, new features are calculated by average-pooling based on these new instance boxes in the same way as DASR~\cite{DASR}. Therefore, the candidate images can be re-ranked with the new similarities from them to the query. \autoref{tab:mAP} shows the search quality of DASR+SPIL in comparison to the instance search performance from representative state-of-the-art methods. In this evaluation, methods based on image-level features~\cite{R-MAC, CroW, BLCF, DeepVision} as well as the ones based on instance-level features~\cite{FCIS+XD, PCL+SPN, DASR} are considered. DASR and our method DASR+SPIL show the best performance. Only minor improvement is observed over DASR, although the produced boxes of our method are much more precise. The localization accuracy no longer makes a significant impact on the discrimination of features when the localization is roughly correct. An offline test confirms that the search quality sees no big difference from ours even the features are pooled with the ground-truth boxes.

\subsection{SPIL on Few-Shot Object Detection}
\label{exp:few-shot-object-detection}

\begin{figure}[tp]
  \centering
  \includegraphics[width=\linewidth]{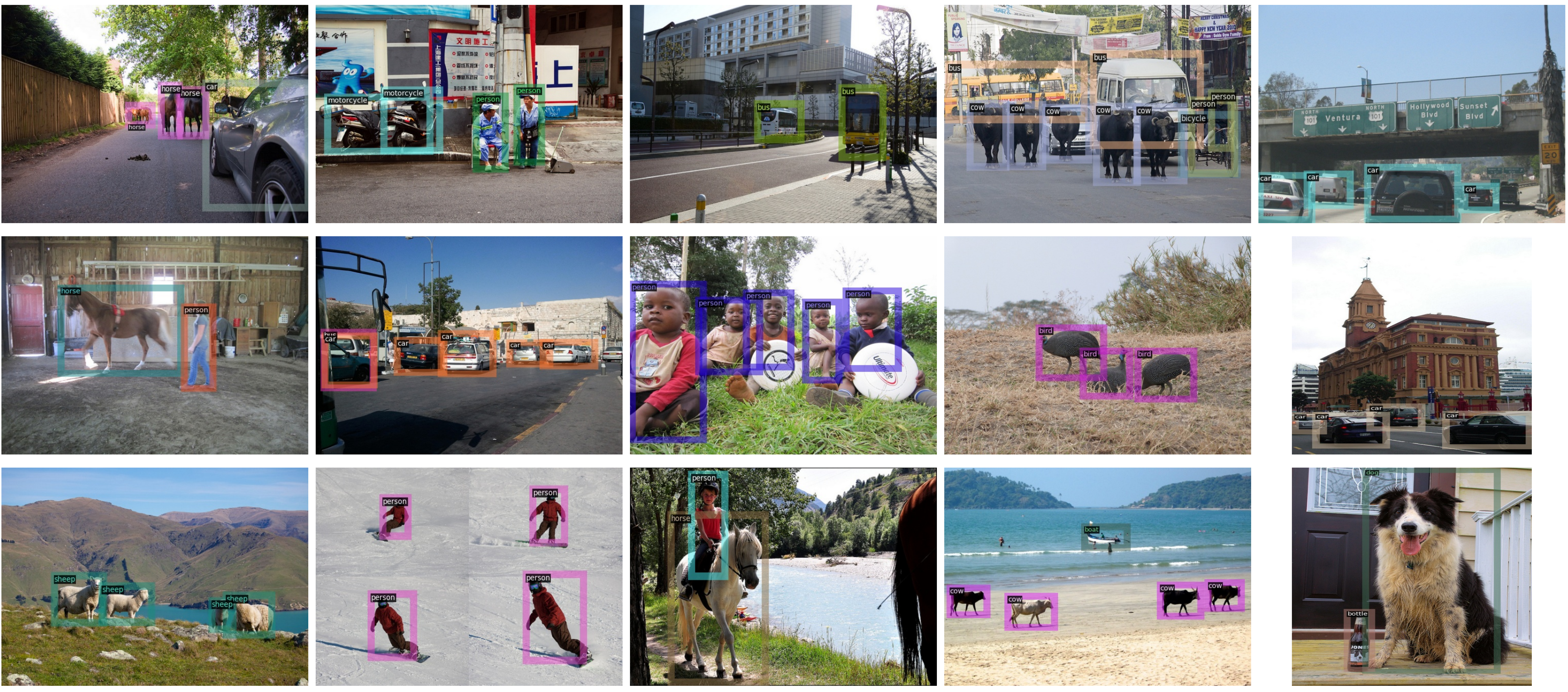}
  \Description{Qualitative 10-shot 20-way object detection results on COCO minival set.}
  \caption{Qualitative 10-shot 20-way object detection results from SPIL on COCO minival set.}
  \label{fig:fsod-example}
\end{figure}

\begin{table}[tb]
  \caption{Few-shot object detection performance on COCO minival set with 10-shot 20-way}
  \label{tab:fsod}
  \begin{tabular}{lcccc}
    \toprule
    Method  & Publication & $AP$ & $AP_{50}$  & $AP_{75}$  \\
    \midrule
    MetaYOLO \cite{FeatureReweighting}  & ICCV 2019 & 5.6 & 12.3  & 4.6 \\ 
    MetaDet     \cite{MetaDet}          & ICCV 2019 & 7.1 & 14.6  & 6.1 \\ 
    Meta R-CNN  \cite{MetaR-CNN}        & ICCV 2019 & 8.7 & 19.1  & 6.6 \\
    TFA* w/cos  \cite{TFA}  & ICML 2020 & 9.1 & 17.1  & 8.8 \\
    Meta-RCNN   \cite{Meta-RCNN}  & ACM MM 2020 & 9.5 & 19.9  & 7.0 \\
    MPSR  \cite{MPSR}   & ECCV 2020 & 9.8   & 17.9  & 9.7 \\
    FSOD  \cite{FewX}   & CVPR 2020 & 11.1  & 20.4  & 10.6 \\
    SRR-FSD \cite{SRR-FSD}  & CVPR 2021 & 11.3  & 23.0  & 9.8 \\
    FSCE \cite{FSCE} & CVPR 2021 & 11.9  & - & 10.5  \\
    \midrule
    SPIL  & - & 11.7  & 23.2 & 10.9 \\
    SPIL+QE & - & \textbf{12.1}  & \textbf{23.6}  & \textbf{11.1} \\
    \bottomrule
  \end{tabular}
\end{table}

In this experiment, the network is trained for \textit{5} stages, \emph{i.e.}, $K=4$. The rest of the hyperparameters are the same as those in the instance search task. In query expansion (QE), we randomly sample up to \textit{10} instances for each query from the pseudo ground-truth instances at Stage-1 since the confidence of the detected instances at this stage is the highest. They are treated as new queries in the second round of instance search. The top-ranked images for these new queries are joined as training images with those from the first round of search.  The performance of method SPIL is compared with state-of-the-art few-shot object detection methods \cite{FeatureReweighting, MetaDet,MetaR-CNN, TFA, Meta-RCNN, MPSR, FewX, SRR-FSD, FSCE}. Experimental results are presented in Table~\ref{tab:fsod}.

As shown in the table, SPIL outperforms most of the state-of-the-art methods. The best performance is achieved when query expansion is integrated. The difficulty for most of the few-shot detection methods lies in the shortage of training data. The available training shots only cover a small part of the feature space. A common practice is to generate prototypes to provide category-related information for novel categories \cite{FeatureReweighting,MetaR-CNN,FSCE,FewX}. However, these prototypes are generated by the few training shots, which cannot cover the diversity of visual objects of one category. Therefore, relying on the few shots to estimate the overall distribution of novel categories inevitably leads to inferior performance. The beauty of SPIL is that it is able to mine more and more reliable training samples by instance search and the afterward query expansion, which could expand the known feature space of one category better than category prototypes. Moreover, our way of detection-by-search relieves the network burden of doing category classification. The network architecture becomes simpler since it only needs to focus on the task of localization. Some qualitative results are shown in Figure \ref{fig:fsod-example}.

%% file: conclusion.tex
\section{Conclusion}
\label{sec:conclusion}
We have presented a novel visual object localization solution named SPIL for both tasks of instance search and few-shot object detection. SPIL aims to achieve accurate object localization using the top-ranked images of instance search. The visual objects can be localized accurately regardless of known or unknown categories when only the box of query object is provided. Compared with the existing few-shot object detection methods, the network is trained with more potential true positives given the fact that a common visual object is shared among top-ranked images and the query. In order to boost the performance, contrastive training and a co-attention module are adopted, both of which in combination lead to considerable performance improvement.

In both instance search and few-shot object detection tasks, SPIL outperforms most of the state-of-the-art approaches. In particular, the localization accuracy in instance search is even higher than strongly supervised methods. Moreover, it maintains high accuracy on the objects of unknown categories. Due to its high localization accuracy, this method is also promising for other downstream tasks such as data-driven object-level image annotation.